\pdfoutput=1

\documentclass[11pt]{article}

\usepackage[final]{acl}

\usepackage{times}
\usepackage{latexsym}

\usepackage[T1]{fontenc}

\usepackage[utf8]{inputenc}

\usepackage{microtype}

\usepackage{inconsolata}

\usepackage{graphicx}
\usepackage{subfig}
\usepackage{longtable}
\usepackage{booktabs} 
\usepackage{multirow} 
\usepackage{siunitx} 
\title{Benchmarking LLMs for Mimicking Child-Caregiver Language in Interaction}

\author{
    Jing Liu$^1$ \quad Abdellah Fourtassi$^2$ \\
    $^1$ENS, PSL Research University, EHESS, CNRS, France \\
    $^2$Aix Marseille Univ, CNRS, LIS, France \\
    \texttt{jing.liu@psl.eu} \quad \texttt{abdellah.fourtassi@gmail.com}
}

\begin{document}
\maketitle
\begin{abstract}
Large Language Models (LLMs) can generate human-like dialogues, yet their ability to simulate early child-adult interactions remains largely unexplored. In this paper, we examined how effectively LLMs can capture the distinctive features of child-caregiver language in interaction, using both static and interactive benchmarking methods. We found that state-of-the-art LLMs like Llama 3 and GPT-4o can approximate child-caregiver dialogues at the word and utterance level, but they struggle to reproduce the child and caregiver's discursive patterns,  exaggerate alignment, and fail to reach the level of diversity shown by humans. The broader goal of this work is to initiate the development of a comprehensive benchmark for LLMs in child-oriented applications.


\end{abstract}

\section{Introduction}

While LLMs show remarkable capabilities in generating human-like text and engaging in open-ended dialogues and role play in various contexts \citep{feng2024can, yang2024talk2care}, their ability to simulate the specificities of child-caregiver interactions has not been systematically investigated. However, these interactions show distinct linguistic and interactive patterns and require dedicated research. 

During their linguistic and communicative development,  children show non-conventional (i.e., non-adult-like) patterns, such as word omissions, mispronunciations, semantic errors, and non-standard grammatical constructions \citep{bloom1993transition}. They also show non-conventional conversational behaviors, such as incoherence, non-responsiveness, and atypical turn-taking patterns \citep{ninio1996pragmatic}. These behaviors are most apparent in the early years through primary school, although many persist into adolescence \citep{nippold2016later}.  

Because of their still immature, non-conventional language use, children depend on caregivers to interpret and clarify their communicative intents, thus facilitating communication. Caregivers employ various \textit{scaffolding} strategies, which offer appropriate support tailored to the child's current level of cognitive and communicative development. These include the general use of simplified language (a register named child-directed language) as well as interactive strategies such as recasting, repairing, providing follow-up, and offering feedback \citep{berk1995scaffolding, clark2020conversational, snow1977talking,nikolaus2023communicative, soderstrom2007beyond}. This scaffolding is gradually reduced as the child becomes more proficient and ready for independent language use.

Despite growing interest in applying LLMs to specialized interactive scenarios \citep{feng2024can, yang2024talk2care}, their ability to simulate child-caregiver interactions remains underexplored. There is, to the best of our knowledge, no systematic examination of whether LLMs can a) properly simulate early child-like utterances with their known non-conventional properties, b) simulate caregiver-like language and its distinctive properties known as child-directed language, and, more importantly, c) beyond mimicking child or caregiver general linguistic properties in \textit{isolation}, simulate child–caregiver language in \textit{interactions}, meaning that we need to simulate the fact that the caregiver's language is responsive/contingent on the child's linguistic quirks, providing tailored feedback and scaffolding. This gap is particularly significant given the potential applications of mimicking this scenario in developmental research and its applications \citep{zhang2024mathemyths, seo2024chacha, rasanen2024age,feng2024can}.

\subsection*{What the current study is about}

As argued above, benchmarking LLMs for effectively mimicking child-caregiver interaction is a complex, multidimensional task. The current work does not claim to provide a complete solution, but rather offers a starting point---an initial exploration into ways to approach some aspect of this task and the insights we learned that can inform future research. 

In particular, it is important to note that a comprehensive benchmarking would require evaluating both the semantic content of interactions (\textit{To what extent was the semantic content of the caregiver-like response appropriate to the child-like utterance?}) and linguistic form (\textit{To what extent was this content linguistically framed in an age-appropriate manner?}). The current work focuses primarily on the latter, using data from spontaneous child-caregiver dialogues as a reference (CHILDES dataset)\cite{macwhinney2000childes}.

Specifically, we quantify the extent to which two widely used LLMs (GPT-4o and Llama 3) approximate child-caregiver interactions along \textit{structural} metrics at word, utterance, and dialogue levels. These metrics were selected based on insights from previous research \citep{valentini2023automatic, rasanen2024age, french2024aligning}. This previous research focused on a few isolated aspects of child-directed dialogue. Here, we aimed to provide a more synthetic view, capitalizing on insights from these studies.

Furthermore, a major novelty in this work is that we compared two benchmarking approaches: single- and multi-turn testing. In single-turn testing (the most common approach), models receive a child utterance from the CHILDES dataset and generate a caregiver-like response, or conversely, receive a caregiver's utterance and generate a child-like response. For multi-turn testing, we observe free interactions, over multiple turns, between a "child"-LLM and a "caregiver"-LLM, prompted to communicate like a child and a caregiver, respectively. The resulting dialogues are then compared, on average, to actual conversations from CHILDES dataset.

Finally, both benchmarking approaches were implemented under zero-shot and few-shot conditions with two objectives: a) for the zero-shot condition, the goal is to characterize the baseline capabilities of LLMs without specific guidance (revealing their initial biases), and b) for the few-shot condition, the goal is to measure LLMs' ability to improve (relative to the zero-shot baseline) when provided with examples and, in fact, test the sensitivity of the metrics we used to mquantify any such improvement.

\section{Related Work and novelty}

Research on LLMs' linguistic and interactive appropriateness to children is still in its early stages. Several studies have explored specific aspects of the interaction, but a comprehensive assessment of LLMs' capabilities in simulating child-caregiver interactions is still underexplored.

For example,  \citet{valentini2023automatic} focused on vocabulary and showed limitations in LLMs' ability to pick simple words for a young audience,
\citet{rasanen2024age} trained a GPT-2 model on caregiver input, and evaluated its capacity to generate language that is similar to caregivers (child-directed language). However, they focused on simulating caregiver data alone, without accounting for the child's data, and therefore, missing the \textit{interactive} dynamics that can influence the generation. \citet{french2024aligning} studied GPT-3.5 and Llama2's ability for linguistic alignment to the interlocutor, showing sub-optimal performance in responding appropriately to child-like utterances. 

While each of these studies has contributed valuable insights to specific dimensions of the challenge, our work aims to provide a more integrated approach on, at least, two levels. First, we evaluate \textit{both} the caregiver- and the child-like generation. We explicitly study them in interaction, not merely simulating them in isolation. Second, instead of focusing on one metric or linguistic level, we propose metrics at the word, utterance, and dialogue levels, aiming to provide a more comprehensive evaluation framework for LLMs in this specialized communicative context.

\section{Method}  

\subsection{Data}  
 
We used the CHILDES public dataset \citep{macwhinney2000childes} for benchmarking, focusing on 2 to 5 years of age.\footnote{This age range was selected because children younger than 2 typically do not engage in extended dialogues, while older age groups had insufficient sample sizes in the dataset to allow for robust analysis.}

From this dataset, we selected 40 conversations (approximately 300 turns each) evenly distributed across the target age groups—specifically, i.e., 10 conversations at 2, 3, 4, and 5 years. 

To prepare the data for analysis and generation, we restructured these conversations into utterance-response pairs.  When consecutive turns came from the same speaker (which is often the case with caregivers), we decomposed these into multiple utterance-response pairs and inserted \texttt{<SILENCE>} tokens to mark the positions where the non-speaking interlocutor did not contribute. This preprocessing step preserved the temporal structure of the interactions while creating a format suitable for our analytical framework. The resulting benchmarking dataset comprised 6,600 interaction pairs containing a total of 73,300 word tokens. This corpus was further characterized by an asymmetric distribution between participant types, with 26,300 tokens produced by children and 47,000 tokens produced by caregivers.

\subsection{Models}  

While numerous LLMs are currently available, testing all of them across multiple experimental conditions would be impractical. Nevertheless, to examine generalizability and ensure that our results are not dependent on the idiosyncrasies of a single model architecture, we systematically compared two state-of-the-art LLMs: an open-source model \textbf{Llama 3 (8B)} (The chat-optimized version of Meta's instruction-tuned large language model, \citep{touvron2023llama}) and a proprietary model; \textbf{GPT-4o}, (version 2024-08),\footnote{The latest version at the moment we wrote this paper.} the generative pre-trained transformer from OpenAI's GPT-4 family \citep{achiam2023gpt}. We selected these models (especially the latest version of ChatGPT) as they are some of the most powerful LLMs (at least at the time this research was carried out) and thus are most likely to be capable of adapting effectively to various roles, including those of children and caregivers.

\subsubsection{Fine-tuned model}  

In addition to the LLMs described above, which we adapt to child-caregiver interaction via prompting, we also evaluated a smaller---and computationally more manageable—pre-trained conventional model that we directly fine-tuned on child-caregiver data. Specifically, we fine-tuned the distilled encoder-decoder BlenderBot model \citep{roller2020recipes} on conversation data from the CHILDES corpus. \footnote{See appendix for fine-tuning details.} This approach allowed us to compare prompt-based adaptation of large, general-purpose models with traditional fine-tuning approaches targeting child-caregiver interaction patterns.

\subsection{Benchmarking}  
\paragraph{Single-turn testing} 

The single-turn testing evaluated LLM's ability to generate contextually appropriate responses to individual utterances. For each conversation in our benchmarking dataset, we implemented two complementary procedures: a) \textbf{Child-to-caregiver direction}: We extracted each child utterance and used it to prompt the LLMs to generate a caregiver-like response. The actual caregiver response from the CHILDES corpus served as the reference against which we evaluated the LLM-generated output. b) \textbf{Caregiver-to-child direction}: Conversely, we extracted each caregiver utterance and used it to prompt the LLMs to generate a child-like response. The actual child response from the corpus served as the reference for evaluation. This single-turn setting allows a direct, controlled comparison between the caregiver-LLMs' response and the caregiver's or child's actual response to the \textit{same} utterance/prompt.

\paragraph{Multi-turn testing}  While single-turn testing enables controlled comparisons with actual responses, it cannot capture the dynamics of extended dialogue interactions. Therefore, we implemented multi-turn testing protocol to evaluate sustained conversational capabilities. To this end, we simulated complete dialogues using two distinct instances of the same LLM: one prompted to behave as a child (hereafter \textit{child-LLM}) and another prompted to behave as a caregiver (hereafter, \textit{caregiver-LLM}). To initiate these simulated conversations, we used the first utterance from each reference conversation in our dataset as a conversation starter. While the outcome is not as controlled as in the single-turn case, the metrics still allow comparisons to the human reference on average, as will be clear next.

\begin{figure*}[t]
\includegraphics[width=1\textwidth]{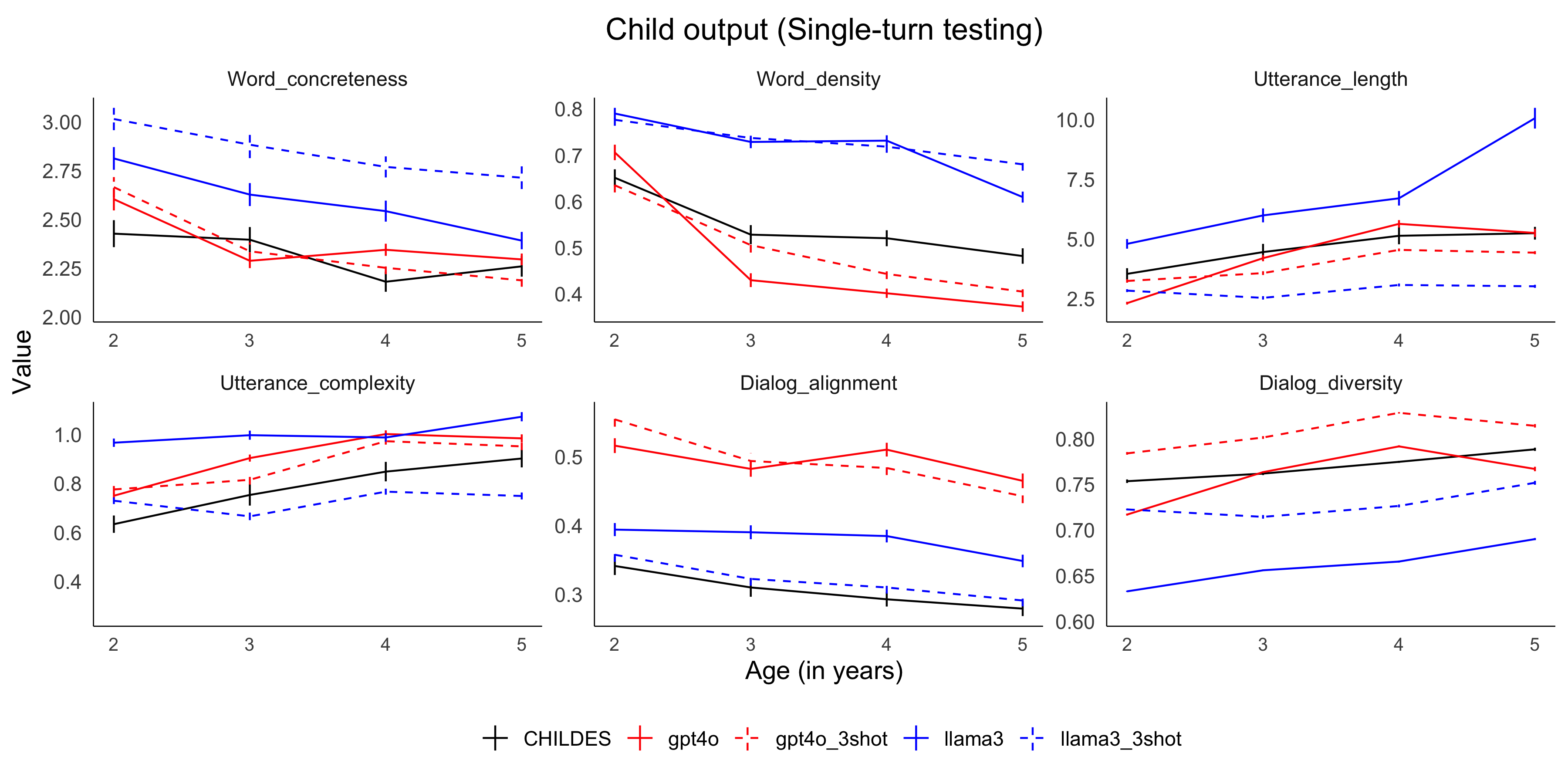}
\caption{Developmental trajectories of six linguistic/conversational features in children and LLMs. We compare actual human data from the child-caregiver dataset CHILDES vs. LLMs playing a child. Models include two LLMs (GPT-4o, Llama 3) in zero-shot and 3-shot settings. Points represent averages across all utterances and conversations, and ranges represent 95\% confidence intervals.}
\label{fig:child_single}
\end{figure*}

\begin{figure*}[t]
\includegraphics[width=1\textwidth]{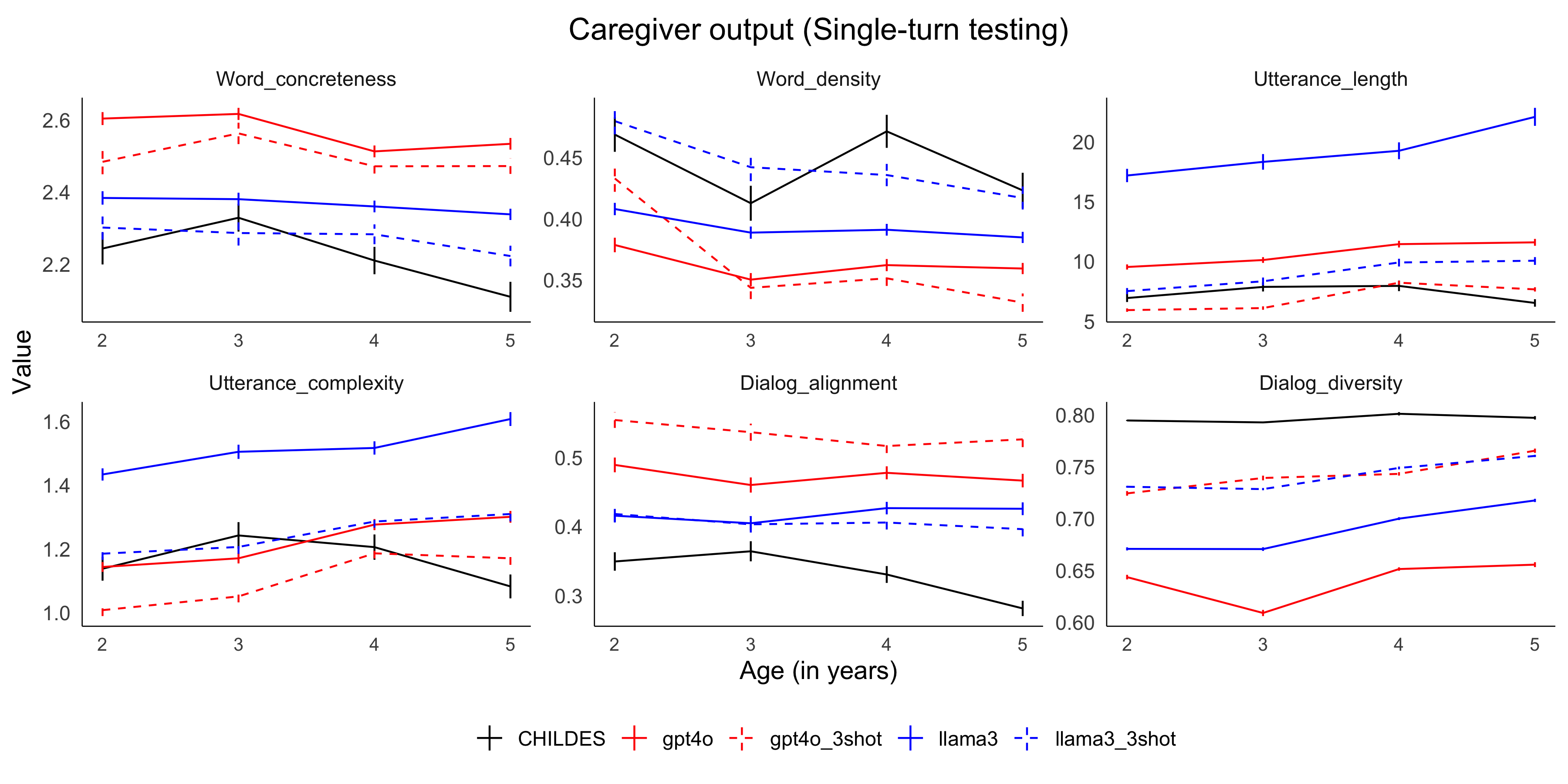}
\caption{Developmental trajectories of six linguistic/conversational features in caregivers and LLMs. We compare actual human data from the child-caregiver dataset CHILDES vs. LLMs playing a caregiver. Models include two LLMs (GPT-4o, Llama 3) in zero-shot and 3-shot settings. Points represent averages across all utterances and conversations, and ranges represent 95\% confidence intervals.}
\label{fig:caregiver_single}
\end{figure*}

\paragraph{Zero vs. Few-shot settings} 

We tested the initial built-in capabilities and biases of the LLMs in a \textbf{zero-shot} setting where LLMs received only the utterance(s) they were expected to respond to, with no additional examples from the CHILDES dataset. In addition, to test our metrics' ability to capture improvement, we tested the same models again in a \textbf{few-shot setting} where the models were given the first three child-caregiver pairs of turns in each conversation, allowing the model to observe examples of both the linguistic structure and interactive dynamics for the same participants. 

In both conditions, we maintained consistent and minimal instructional prompts to avoid introducing confounding variables. For example, when prompting for caregiver responses, we used instructions such as: ``\textit{You are the parent of a X-year-old English-speaking child. Now, you are having a conversation
with your child. Based on the conversation history above, give your response to the child input.}'') (see the prompt templates in Appendix~\ref{sec:prompts}).

\subsection{Metrics}

We aim to provide a comprehensive benchmarking evaluating key properties at the word-, utterance-, and dialogue levels.

At the \textit{word} level, we followed \citet{dawson2021features} to quantify the \textit{Word concreteness} using human ratings from \citet{brysbaert2014concreteness}. This measure was computed as the average concreteness rating of all content words in each utterance. We also used \textit{Word density}---defined as the proportion of content (vs. function) words in the utterance in each utterance. This metric reflects the information load and was calculated using the established list of function words by \citet{o2012multi}.

At the \textbf{utterance} level, we adopted measures from \citet{rasanen2024age} to capture structural complexity: a) \textit{Utterance length}: the number of words per utterance; b) \textit{Syntactic complexity}: the mean dependency tree depth for each utterance using the Spacy toolkit (v3.7) with its dependency parser based on RoBERTa transformers.\footnote{The dependency parsing implementation provides a structural representation of syntactic relationships between words in an utterance.} \citep{liu2008dependency}, where deeper trees indicate more complex structures. 

Finally, at the level of the \textbf{dialogue} dynamics, we measured \textit{Semantic alignment}, the extent to which the speaker's utterance is semantically similar to their interlocutor's (across each exchange pair) \citep{duran2019align, french2024aligning, misiek2020development}. To this end, we used BERT sentence embedding \citep{reimers-2019-sentence-bert}. Following \citet{guo2023curious}, this was calculated as the average pairwise cosine distance (1 - cosine similarity) between BERT utterance embeddings of a speaker's contributions across the entire conversation.

\section{Results and Discussion}

\begin{figure*}[t]
\includegraphics[width=1\textwidth]{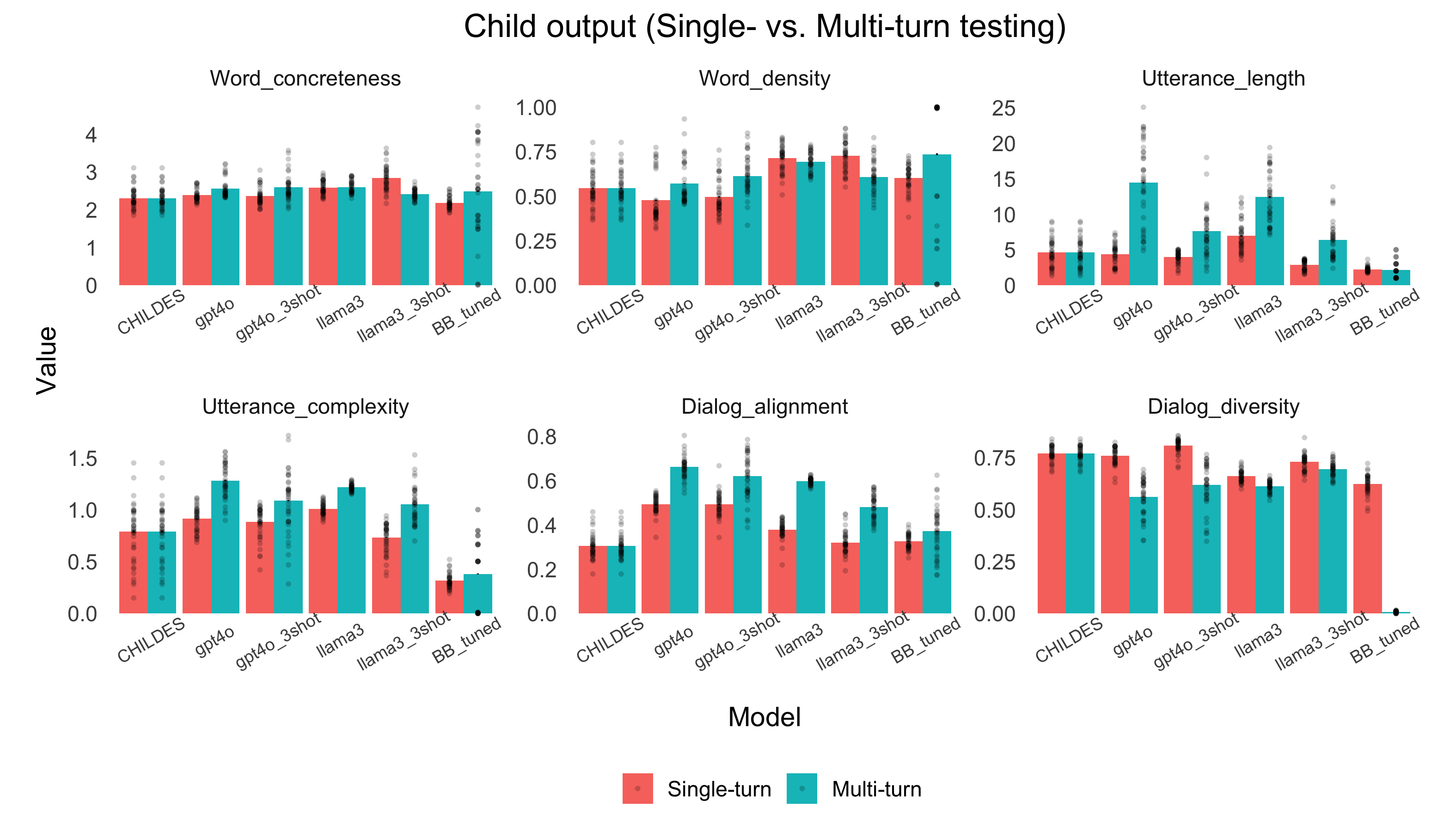}
\caption{Comparison of linguistic/conversational features across the single-turn and multi-turn testing for children. The bars show the averages across age groups. The points show variability across conversations (40 in total). Models include two LLMs (GPT-4o, Llama 3) in zero-shot and 3-shot settings. ``BB\_tuned'' which stands for the BlenderBot model fine-tuned on a subset of CHILDES. Finally, for CHILDES, the data for single-turn and multi-turn is the same.}
\label{fig:child_multi}
\end{figure*}

\begin{figure*}[ht]
\includegraphics[width=1\textwidth]{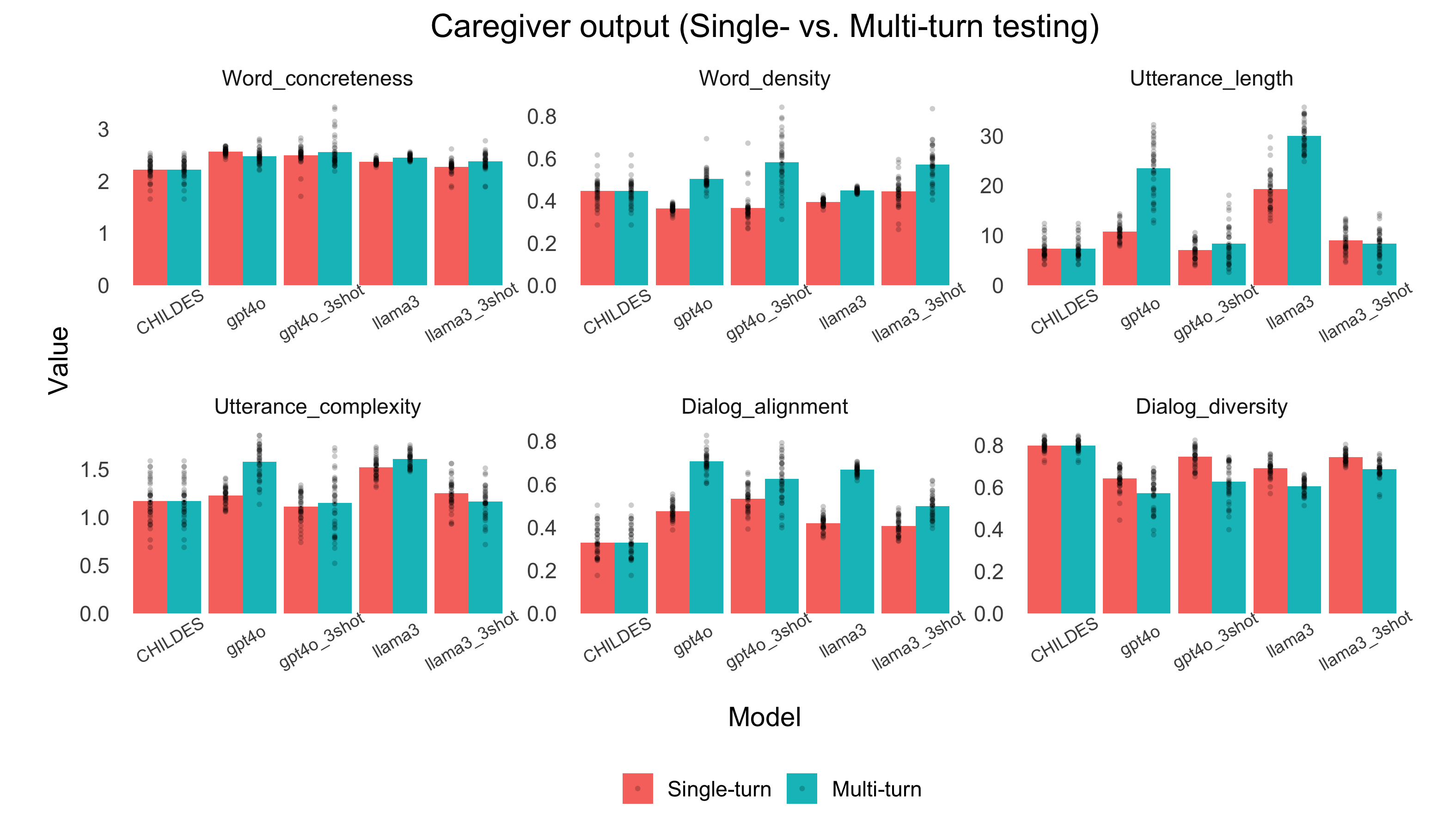}
\caption{Comparison of linguistic/conversational features across the single-turn and multi-turn testing for caregivers. The bars show the averages across age groups. The points show variability across conversations (40 in total). Models include two LLMs (GPT-4o, Llama 3) in zero-shot and 3-shot settings. Finally, for CHILDES, the data for single-turn and multi-turn is the same.}
\label{fig:caregiver_multi}
\end{figure*}

\subsection{Single-turn testing}

Figure \ref{fig:child_single} and Figure \ref{fig:caregiver_single} show the results of single-turn testing for child-LLMs and caregiver-LLMs across age groups.

\paragraph{Child-LLM} 

Our analysis reveals that child-LLMs follow, overall, the developmental patterns observed in CHILDES, even in the zero-shot setting, showing that these LLMs can simulate developmental changes in children's language without explicit guidance.  

Specifically, the models capture decreasing concreteness and lexical density over time (children use more abstract words and function words as they grow older), increasing utterance length and syntactic complexity, and decreasing dialog alignment alongside increasing dialog diversity (as children become able to contribute new information instead of just repeating caregivers' inputs).

When comparing model performance, GPT-4o more accurately mimicked children than Llama 3, achieving closer alignment with CHILDES reference values across most metrics in the zero-shot condition. The introduction of few-shot examples (three interaction pairs) primarily benefited Llama 3, improving its alignment with CHILDES across several measures, though not consistently across all metrics. Despite these improvements, GPT-4o maintained better overall performance following few-shot prompting.

\paragraph{Caregiver-LLM} In simulating caregiver language, zero-shot GPT-4o more closely approximated CHILDES values for utterance-level measures, while Llama 3 achieved better alignment with word-level and dialogue-level metrics. Few-shot prompting significantly improved both models' performance, with Llama 3 ultimately achieving better overall alignment with CHILDES reference values than GPT-4o after exposure to exemplars. Notably, both models continued to diverge from human data at the dialogue level, exhibiting higher semantic alignment and lower semantic diversity than actual caregivers in the CHILDES corpus. Overall, there were no clear developmental patterns in CHILDES caregivers' behaviors.

\subsection{Multi-turn testing}

Figure \ref{fig:child_multi} and \ref{fig:caregiver_multi} show the results of multi-turn testing. For easier comparison with single-turn, we show the results side-by-side, averaged over age. Here, we made a caregiver-LLM interact with a child-LLM (using two instances of the same LLM).\footnote{We also crossed models in child-LLM vs. caregiver-LLM (e.g., child-llama 3 interacting with caregiver-GPT-4o), but observed no noticeable changes.}

\paragraph{Child-LLM} In zero-shot settings, we observed marked differences between single-turn and multi-turn testing across multiple metrics. Most notably, utterance length, syntactic complexity, and semantic alignment all increased considerably, becoming less comparable to children in CHILDES.

After few-shot prompting, we observed minor to moderate improvements, such as reductions in utterance length \footnote{Note that length could not be reduced further, even when we experimented with explicitly setting an upper bound on children's utterance in our instructions to the LLM.} and (slight) decreases in both syntactic complexity and semantic alignment. \footnote{In addition to using LLMs instructed to play a child, we also used model fine-tuning. To this end, we considered a pre-trained encoder-decoder model- Blenderbot fine-tuned on a subset of caregiver(encoder)-child(decoder) dialogues in CHILDES. We also explored an alternative approach using model fine-tuning rather than prompting. Specifically, we fine-tuned a pre-trained encoder-decoder Blenderbot model on a subset of caregiver (encoder) to child (decoder) dialogues from CHILDES. However, while showing reasonable performance in single-turn testing, this fine-tuned model was erratic and highly repetitive in the multi-turn interactions, also making it unsuitable for evaluating caregiver-LLMs.}

\paragraph{Caregiver-LLM} Multi-turn testing resulted in a general increase across most metrics compared to single-turn testing (with exceptions in dialog diversity, which decreased, and concreteness, which remained constant). These changes made the multi-turn behavior generally less comparable to human data in CHILDES. After the few-shot learning, we observed a significant improvement in multi-turn behavior, especially in terms of length and complexity, which became much more comparable to CHILDES. However, we observed only moderate improvement in dialog-level measures.

While LLMs successfully approximated human references for word-level and sentence-level properties under certain conditions, we identified systematic discrepancies in interactive measures for caregiver-LLMs across all experimental configurations. As illustrated in Figure \ref{fig:caregiver_multi}, in both dialog alignment and diversity, LLMs exhibited higher alignment and lower diversity than caregiver data in CHILDES across LLM type (GPT-4o  LLama 3), prompting strategy (zero- vs few-shot) and benchmarking approach (Single- vs. multi-turn).

To verify this qualitative observation statistically, we ran, for each of the two interactive measures (Dialig\_alignment and Dialog\_diversity), linear regressions comparing models' output to the CHILDES reference, testing all configurations: 2 LLMs x 2 promoting strategy x 2 benchmarking approach. All 16 comparison models (8 for each measure) revealed highly statistically significant differences between LLM-generated and human caregiver language, suggesting that interactive caregiver properties are potentially more challenging to mimic for LLM.

\section{Discussion}

This paper presents a preliminary exploration into ways we can benchmark LLMs' ability to simulate child-caregiver interactive dynamics. While previous research has typically focused on measuring the properties of a specific structure/level (e.g., words or utterances) and/or on evaluating language from one part of the dialogue outside the interactive context (mainly focusing on the caregiver), our novel contribution is in three key aspects: a) we evaluate child-caregvier language generation in \textit{interaction},  b) we tested a more comprehensive set of measures covering the word, utterance, and dialog levels, and c) we compared two benchmarking approaches, evaluating the models in short and extended settings. 


We put this benchmarking framework to use, comparing two powerful LLMs: GPT-4 and LLama3. Indeed, such a comparison was essential to distinguish between findings that are likely generalizable and those that are specific to a particular model. Our evaluation incorporated both single- and multi-turn testing. The former is more controlled, allowing a direct evaluation of the LLMs' response using human references, while the latter, though less controlled, allowed measuring LLMs' behavior in more extended conversations. Interestingly, we found that single-turn evaluations, while insightful, were not totally correlated with the LLMs' behavior in an extended conversation. Indeed, the multi-turn analyses showed an increased divergence of LLMs from human data, particularly in utterance-level properties and discourse dynamics (although both benchmarkings led to qualitatively similar conclusions regarding this linguistic level). This quantitative mismatch highlights the importance of dynamic testing of LLMs, since static testing alone may fail to capture the cumulative effects of sustained interactions---a consideration particularly relevant for applications involving extended communicative exchanges.

Our benchmarking effort aims not only at comparing different LLMs, but also at providing a quantitative tool that can be sensitive to incremental improvement \textit{within} the same models. In particular, comparing zero- vs. few-shot learning, the benchmark identified areas where this intervention was more or less successful: Both GPT-4o and Llama 3 showed marked improvements in matching caregiver patterns after exposure to just three interactive examples---a finding with important implications for efficient model adaptation in resource-constrained contexts. 

However, the impact was consistently less pronounced for dialogue-level properties, where models (whether in single- or multi-turn settings) continued to exhibit higher alignment and lower diversity compared to human data. This pattern suggests that while surface-level linguistic features (such as word and sentence properties) appear relatively straightforward to adapt to, capturing the interactive nature of child-caregiver communication is more challenging (see also Limitations).

In conclusion, we explored a multi-level benchmarking approach for assessing LLMs' ability to mimic child-caregiver language in interactions, and we showed its usefulness in tracking incremental improvements. We found that single-turn testing of the LLMs, as typical in most benchmarks, was not totally indicative of the LLMs' real behavior in extended conversational contexts, thus emphasizing the need for more dynamic, multi-turn testing in this line of work. Furthermore, few-shot prompting was effective in bringing the LLMs closer to caregivers' data, especially regarding word- and utterance-level properties. It was not as effective on the dialog-level properties. The LLMs exaggerated alignment and showed reduced diversity compared to CHILDES.

\section{Limitations}

While our explorative work has led to some initial insights, it also has several limitations and raises questions for future research.

A primary constraint concerns our selection of evaluation metrics. We relied on established measures from previous research to assess LLMs at word, sentence, and dialogue levels. For children's language production, these metrics successfully captured developmental trajectories, enabling age-specific benchmarking of LLMs (Figure 1). However, when applied to caregiver language, these same metrics failed to reveal clear developmental patterns (Figure 2), thereby limiting the precision of our assessment. This asymmetry highlights the need for developing more refined metrics that can better capture the subtle adaptations in caregiver speech across different stages of child development. 

While evaluating LLMs in the single-turn scenario---comparing answers to a reference---is rather straightforward and aligns with the way LLMs are typically benchmarked, a multi-turn approach is also necessary in our context since the ultimate goal behind benchmarking is for these models to be used in an extended interactive context. However, the multi-turn approach is inherently less controlled as it involves two instances of LLMs interacting, each playing the role of an interlocutor. However, since neither instance perfectly models the interlocutor, extended interaction can amplify artifacts, leading the systems to adapt to each other’s quirks rather than approximating genuine child–caregiver interaction. Thus, a multi-turn benchmarking approach can under-estimate the models' true capabilities to interact with an actual human (child or caregiver). That said, the fact that multi-turn testing was, at least qualitatively, in agreement with single-turn testing (e.g., both of them point to LLMs' ability to adapt more easily to properties of words and sentences and less easily to properties of the dialog) is a testimony to the multi-turn's potential as a valid method of evaluation in this context, though, of course, future research is needed to thoroughly investigate this question.

Finally, we reported that few-shot learning improved caregiver-LLM performance in terms of utterance length and syntactic complexity, but had less impact on alignment and diversity. While this finding was, overall, robust across configurations in our setup, it should be taken with a grain of salt, given that we did not systematically test it in a comprehensive set of experiments. Indeed, our main goal in this paper, as we stated above, was not to improve the LLMs' performance, but to demonstrate, in a simple case, the ability of the benchmarking metrics to identify gaps and track improvement. For example, it is possible that a larger, more systematic exploration of the prompting strategies (which is computationally expensive, and thus, could not be done here in combination with the other experiments we did) could yield improvements across the dialog dimensions as well.

\section{Ethics statement}
All data used in this study is already publicly available.
This work focuses on model benchmarking and improvement using offline child-caregiver data and internal simulations, aiming to advance fundamental research in this area. We do not consider this testing sufficient for deployment; any future real-life applications should undergo rigorous validation in child-safe environments with appropriate human oversight, such as by teachers or parents.


\section*{Acknowledgments}
This project has received funding from the European Union's Horizon 2020 research and innovation programme under the Marie Skłodowska-Curie grant agreement No 945304 – Cofund AI4theSciences hosted by PSL University.

\bibliography{custom}

\begin{thebibliography}{27}
\providecommand{\natexlab}[1]{#1}

\bibitem[{Achiam et~al.(2023)Achiam, Adler, Agarwal, Ahmad, Akkaya, Aleman, Almeida, Altenschmidt, Altman, Anadkat et~al.}]{achiam2023gpt}
Josh Achiam, Steven Adler, Sandhini Agarwal, Lama Ahmad, Ilge Akkaya, Florencia~Leoni Aleman, Diogo Almeida, Janko Altenschmidt, Sam Altman, Shyamal Anadkat, et~al. 2023.
\newblock Gpt-4 technical report.
\newblock \emph{arXiv preprint arXiv:2303.08774}.

\bibitem[{Berk and Winsler(1995)}]{berk1995scaffolding}
Laura~E Berk and Adam Winsler. 1995.
\newblock \emph{Scaffolding Children's Learning: Vygotsky and Early Childhood Education. NAEYC Research into Practice Series. Volume 7.}
\newblock ERIC.

\bibitem[{Bloom(1993)}]{bloom1993transition}
L~Bloom. 1993.
\newblock \emph{The transition from infancy to language: Acquiring the power of expression}.
\newblock Cambridge University Press.

\bibitem[{Brysbaert et~al.(2014)Brysbaert, Warriner, and Kuperman}]{brysbaert2014concreteness}
Marc Brysbaert, Amy~Beth Warriner, and Victor Kuperman. 2014.
\newblock Concreteness ratings for 40 thousand generally known english word lemmas.
\newblock \emph{Behavior research methods}, 46:904--911.

\bibitem[{Clark(2020)}]{clark2020conversational}
Eve~V Clark. 2020.
\newblock Conversational repair and the acquisition of language.
\newblock \emph{Discourse Processes}, 57(5-6):441--459.

\bibitem[{Dawson et~al.(2021)Dawson, Hsiao, Banerji, Tan, and Nation}]{dawson2021features}
Nicola Dawson, Yaling Hsiao, N~Banerji, AWM Tan, and Kate Nation. 2021.
\newblock Features of lexical richness in children’s books: Comparisons with child-directed speech.
\newblock \emph{Language Development Research}.

\bibitem[{Duran et~al.(2019)Duran, Paxton, and Fusaroli}]{duran2019align}
Nicholas~D Duran, Alexandra Paxton, and Riccardo Fusaroli. 2019.
\newblock Align: Analyzing linguistic interactions with generalizable techniques—a python library.
\newblock \emph{Psychological methods}, 24(4):419.

\bibitem[{Feng et~al.(2024)Feng, Xu, Lahiri, Tager-Flusberg, Kim, Bishop, Lord, and Narayanan}]{feng2024can}
Tiantian Feng, Anfeng Xu, Rimita Lahiri, Helen Tager-Flusberg, So~Hyun Kim, Somer Bishop, Catherine Lord, and Shrikanth Narayanan. 2024.
\newblock Can generic llms help analyze child-adult interactions involving children with autism in clinical observation?
\newblock \emph{arXiv preprint arXiv:2411.10761}.

\bibitem[{French et~al.(2024)French, D’Mello, and Wense}]{french2024aligning}
Dorothea French, Sidney D’Mello, and Katharina Wense. 2024.
\newblock Aligning to adults is easy, aligning to children is hard: A study of linguistic alignment in dialogue systems.
\newblock In \emph{Proceedings of the 1st Human-Centered Large Language Modeling Workshop}, pages 81--87.

\bibitem[{Guo et~al.(2023)Guo, Shang, Vazirgiannis, and Clavel}]{guo2023curious}
Yanzhu Guo, Guokan Shang, Michalis Vazirgiannis, and Chlo{\'e} Clavel. 2023.
\newblock The curious decline of linguistic diversity: Training language models on synthetic text.
\newblock \emph{arXiv preprint arXiv:2311.09807}.

\bibitem[{Liu(2008)}]{liu2008dependency}
Haitao Liu. 2008.
\newblock Dependency distance as a metric of language comprehension difficulty.
\newblock \emph{Journal of Cognitive Science}, 9(2):159--191.

\bibitem[{MacWhinney(2000)}]{macwhinney2000childes}
Brian MacWhinney. 2000.
\newblock The childes project.
\newblock \emph{Computational Linguistics}, 26(4):657--657.

\bibitem[{Misiek et~al.(2020)Misiek, Favre, and Fourtassi}]{misiek2020development}
Thomas Misiek, Benoit Favre, and Abdellah Fourtassi. 2020.
\newblock Development of multi-level linguistic alignment in child-adult conversations.
\newblock In \emph{Proceedings of the workshop on cognitive modeling and computational linguistics}, pages 54--58.

\bibitem[{Nikolaus and Fourtassi(2023)}]{nikolaus2023communicative}
Mitja Nikolaus and Abdellah Fourtassi. 2023.
\newblock Communicative feedback in language acquisition.
\newblock \emph{New Ideas in Psychology}, 68:100985.

\bibitem[{Ninio and Snow(1996)}]{ninio1996pragmatic}
Anat Ninio and Catherine~E. Snow. 1996.
\newblock \emph{Pragmatic Development}.
\newblock Westview Press.

\bibitem[{Nippold(2016)}]{nippold2016later}
Marilyn~A Nippold. 2016.
\newblock \emph{Later language development: School-age children, adolescents, and young adults}.
\newblock ERIC.

\bibitem[{O’Shea et~al.(2012)O’Shea, Bandar, and Crockett}]{o2012multi}
James O’Shea, Zuhair Bandar, and Keeley Crockett. 2012.
\newblock A multi-classifier approach to dialogue act classification using function words.
\newblock \emph{Transactions on Computational Collective Intelligence VII}, pages 119--143.

\bibitem[{R{\"a}s{\"a}nen and Kocharov(2024)}]{rasanen2024age}
Okko R{\"a}s{\"a}nen and Daniil Kocharov. 2024.
\newblock Age-dependent analysis and stochastic generation of child-directed speech.
\newblock \emph{arXiv preprint arXiv:2405.07700}.

\bibitem[{Reimers and Gurevych(2019)}]{reimers-2019-sentence-bert}
Nils Reimers and Iryna Gurevych. 2019.
\newblock \href {http://arxiv.org/abs/1908.10084} {Sentence-bert: Sentence embeddings using siamese bert-networks}.
\newblock In \emph{Proceedings of the 2019 Conference on Empirical Methods in Natural Language Processing}. Association for Computational Linguistics.

\bibitem[{Roller(2020)}]{roller2020recipes}
S~Roller. 2020.
\newblock Recipes for building an open-domain chatbot.
\newblock \emph{arXiv preprint arXiv:2004.13637}.

\bibitem[{Seo et~al.(2024)Seo, Yang, and Kim}]{seo2024chacha}
Woosuk Seo, Chanmo Yang, and Young-Ho Kim. 2024.
\newblock Chacha: Leveraging large language models to prompt children to share their emotions about personal events.
\newblock In \emph{Proceedings of the CHI Conference on Human Factors in Computing Systems}, pages 1--20.

\bibitem[{Snow(1977)}]{snow1977talking}
CE~Snow. 1977.
\newblock Talking to children: Language input and acquisition.

\bibitem[{Soderstrom(2007)}]{soderstrom2007beyond}
Melanie Soderstrom. 2007.
\newblock Beyond babytalk: Re-evaluating the nature and content of speech input to preverbal infants.
\newblock \emph{Developmental Review}, 27(4):501--532.

\bibitem[{Touvron et~al.(2023)Touvron, Lavril, Izacard, Martinet, Lachaux, Lacroix, Rozi{\`e}re, Goyal, Hambro, Azhar et~al.}]{touvron2023llama}
Hugo Touvron, Thibaut Lavril, Gautier Izacard, Xavier Martinet, Marie-Anne Lachaux, Timoth{\'e}e Lacroix, Baptiste Rozi{\`e}re, Naman Goyal, Eric Hambro, Faisal Azhar, et~al. 2023.
\newblock Llama: Open and efficient foundation language models.
\newblock \emph{arXiv preprint arXiv:2302.13971}.

\bibitem[{Valentini et~al.(2023)Valentini, Weber, Salcido, Wright, Colunga, and Kann}]{valentini2023automatic}
Maria Valentini, Jennifer Weber, Jesus Salcido, T{\'e}a Wright, Eliana Colunga, and Katharina Kann. 2023.
\newblock On the automatic generation and simplification of children's stories.
\newblock \emph{arXiv preprint arXiv:2310.18502}.

\bibitem[{Yang et~al.(2024)Yang, Xu, Yao, Rogers, Zhang, Intille, Shara, Gao, and Wang}]{yang2024talk2care}
Ziqi Yang, Xuhai Xu, Bingsheng Yao, Ethan Rogers, Shao Zhang, Stephen Intille, Nawar Shara, Guodong~Gordon Gao, and Dakuo Wang. 2024.
\newblock Talk2care: An llm-based voice assistant for communication between healthcare providers and older adults.
\newblock \emph{Proceedings of the ACM on Interactive, Mobile, Wearable and Ubiquitous Technologies}, 8(2):1--35.

\bibitem[{Zhang et~al.(2024)Zhang, Liu, Ziska, Jeon, Yu, and Xu}]{zhang2024mathemyths}
Chao Zhang, Xuechen Liu, Katherine Ziska, Soobin Jeon, Chi-Lin Yu, and Ying Xu. 2024.
\newblock Mathemyths: leveraging large language models to teach mathematical language through child-ai co-creative storytelling.
\newblock In \emph{Proceedings of the CHI Conference on Human Factors in Computing Systems}, pages 1--23.

\end{thebibliography}

\onecolumn

\setlength{\parskip}{6pt}

\clearpage
\appendix

\section*{Appendix A: Prompt templates}
\label{sec:prompts}
We provide prompt templates in different conditions. These are templates for the multi-turn testing. The templates for the single-turn testing are the same, except there is no conversation history.

\paragraph{The zero-shot prompt template for the caregiver}

[\textbf{Conversation history} ADULT: \textless Utterance\textgreater, CHI: \textless Utterance\textgreater ...]
You are the parent of a \textless Month\textgreater-month-old English-speaking child. Now, you are having a conversation with your child. \texttt{<SILENCE>} indicates silence in the previous turn; \texttt{<UNINTELLIGIBLE>} indicates unintelligible speech. Based on the given conversation history above, give your response to parent input as ADULT. Do not output the speaker label.

\paragraph{The zero-shot prompt template for the child}

[\textbf{Conversation history} CHI: \textless Utterance\textgreater, ADULT: \textless Utterance\textgreater ...]
You are a \textless Month\textgreater-month-old English-speaking child. Now, you are having a conversation with your parent. \texttt{<SILENCE>} indicates silence in the previous turn; \texttt{<UNINTELLIGIBLE>} indicates unintelligible speech. Based on the given conversation history above, give your response to parent input as CHI. Do not output the speaker label.

\paragraph{The few-shot prompt template for the caregiver}

[\textbf{Conversation history} ADULT: \textless Utterance\textgreater, CHI: \textless Utterance\textgreater ...]
You are the parent of a \textless Month\textgreater-month-old English-speaking child. Now, you are having a conversation with your child. \texttt{<SILENCE>} indicates silence in the previous turn; \texttt{<UNINTELLIGIBLE>} indicates unintelligible speech. Ensure your response is no longer than 50 words regardless of the prompt. Here are some example interactions: CHI: \textless Utterance\textgreater, ADULT: \textless Utterance\textgreater ... Follow the example interactions. Based on the given conversation history above, give your response to parent input as ADULT. Do not output the speaker label.

\paragraph{The few-shot prompt template for the child}

[\textbf{Conversation history} CHI: \textless Utterance\textgreater, ADULT: \textless Utterance\textgreater ...]
You are a \textless Month\textgreater-month-old English-speaking child. Now, you are having a conversation with your parent. \texttt{<SILENCE>} indicates silence in the previous turn; \texttt{<UNINTELLIGIBLE>} indicates unintelligible speech. Ensure your response is no longer than 6 words regardless of the prompt. Here are some example interactions: CHI: \textless Utterance\textgreater, ADULT: \textless Utterance\textgreater ... Follow the example interactions. Based on the given conversation history above, give your response to parent input as CHI. Do not output the speaker label.

\section*{Appendix B: Finetuning details}
We fine-tuned the Blenderbot model, a 400M parameter architecture comprising: a retriever for dialogue history; a Seq2Seq generator with 2 encoder layers, 24 decoder layers, 2560-dimensional embeddings, and 32 attention heads; retrieve-and-refine architectures combining dialogue and knowledge retrieval. The parameters were optimized through preliminary experiments, resulting in a batch size of 16 and a learning rate of 0.0001 maintained over 4 epochs. The fine-tuning data consisted of 2.6M interaction pairs with 15.8M total word tokens (4.4M child tokens and 11.4M adult tokens) restructured from CHILDES dataset \citep{macwhinney2000childes}.

\section*{Appendix C: Example generations}

We present sample generations in Table~\ref{tab:samples}. It shows child-caregiver conversations (CHILDES corpus) and GPT-4o generations in a multi-turn setting where the child and caregiver are two instances of GPT-4o. All conversations start from the same child turn (here, all conversations start from a child outputting ``silence''). To extract features, we preprocessed all samples by removing non-verbal behavioral annotations (e.g., \textit{giggles}) and lower-casing words to focus exclusively on verbal content. 

\begin{table*}[h]
    \small
    \centering
    \caption{Comparison of real and LLM-generated child-caregiver dialogue exchanges. The LLM generation here is from a multi-turn setting, where two instances of GPT-4o (i.e., a child-LLM and a caregiver-LLM) interact. The table shows three dialogue samples: (1) a real interaction from the CHILDES corpus, (2) a zero-shot generation by GPT-4o given only the initial turn, and (3) a few-shot generation by GPT-4o after being shown three previous examples in the child-caregiver interaction.}
    \label{tab:samples}
    \setlength{\tabcolsep}{4pt}
    \begin{tabular}{l l p{0.8\textwidth}}
    \toprule
    \textbf{Source} & \textbf{Role} & \textbf{Utterance} \\
    \midrule
    \multirow{8}{*}{\shortstack[l]{CHILDES\\(Human)}} 
    & C & \textit{SILENCE} \\
    & A & wanna play catch \\
    & C & let's play catch \\
    & A & okay almost \\
    & C & yeah \\
    & A & try it again put your hands in like this \\
    & C & almost \\
    & A & you wanna hit it \\
    \midrule
    \multirow{8}{*}{\shortstack[l]{GPT-4o\\(Zero-shot)}} 
    & C & \textit{SILENCE} \\
    & A & I see you're still thinking about cherries. It's okay to feel upset. Would you like a hug while we figure out something else yummy to eat? \\
    & C & \textit{SILENCE} \\
    & A & I know you're feeling quiet right now. I'm here when you're ready. Would you like to hold the banana and see if it feels fun to peel? \\
    & C & Banana! Peel it? \\
    & A & Yes, let's peel the banana! Can you help me pull the peel down? Then we can eat it together. Yum! \\
    & C & Banana! Peel! Eat! \\
    & A & Yes! Let’s peel the banana together. Peel, peel! Now take a bite—yum! Is it tasty? \\
    \midrule
    \multirow{8}{*}{\shortstack[l]{GPT-4o\\(Few-shot)}} 
    & C & \textit{SILENCE} \\
    & A & Do you love your big red truck? \\
    & C & Big red truck! My truck! \\
    & A & Is it your big red truck? Where will you drive it? \\
    & C & Drive to park! Big red truck! \\
    & A & Are we driving your big red truck to the park? That sounds so fun! \\
    & C & My truck drive park! \\
    & A & Yes, your big red truck! We can drive it to the park! \\
    
    \bottomrule
    \end{tabular}
    
    \vspace{1mm}
    \raggedright
    \footnotesize
    \textbf{Notes:} A = Adult, C = Child. All model-generated responses were obtained using default sampling parameters (temperature = 1.0). \textit{SILENCE} marks an empty turn. Human dialogues are from the CHILDES corpus \citep{macwhinney2000childes}.
\end{table*}

\end{document}